\documentclass[letterpaper, 10 pt, conference]{ieeeconf}  

\IEEEoverridecommandlockouts                              

\usepackage{graphics} 
\usepackage{epsfig} 
\usepackage{times} 
\usepackage{amsmath} 
\usepackage{amssymb}  
\usepackage{centernot}
\usepackage{pbox}
\usepackage{svg}
\usepackage{bm}

\usepackage{url}
\usepackage{hyperref}

\usepackage{algorithm}
\usepackage[noend]{algpseudocode}
\algnewcommand\algorithmicforeach{\textbf{for each}}
\algdef{S}[FOR]{ForEach}[1]{\algorithmicforeach\  #1\ \algorithmicdo}

\usepackage{booktabs}

\usepackage{xcolor}

\DeclareMathOperator*{\argmin}{\arg\!\min}

\newcommand\methodname{Cross Entropy Gait Optimization for Legged Systems}
\newcommand\acronym{CrEGOpt} 

\title{\LARGE \bf
Gait Optimization for Legged Systems Through Mixed Distribution Cross-Entropy Optimization
}

\author{Ioannis Tsikelis$^{1\dagger}$ and Konstantinos Chatzilygeroudis$^{2,3\dagger}$
\thanks{$^{\dagger}$Equal Contribution}%
\thanks{$^{1}$Computer Engineering and Informatics Department,
        University of Patras, GR-26504 Patras, Greece,
        {\tt\small tsikelis.i@protonmail.com}}%
\thanks{$^{2}$Laboratory of Automation and Robotics (LAR) in the Department of Electrical \& Computer Engineering,
        University of Patras, GR-26504 Patras, Greece,
        {\tt\small costashatz@upatras.gr}}%
\thanks{$^{3}$Computational Intelligence Laboratory (CILab), Department of Mathematics,
        University of Patras, GR-26110 Patras, Greece}%
}

\begin{document}

\maketitle
\thispagestyle{empty}
\pagestyle{empty}

\begin{abstract}
Legged robotic systems can play an important role in real-world applications due to their superior load-bearing capabilities, enhanced autonomy, and effective navigation on uneven terrain. They offer an optimal trade-off between mobility and payload capacity, excelling in diverse environments while maintaining efficiency in transporting heavy loads. However, planning and optimizing gaits and gait sequences for these robots presents significant challenges due to the complexity of their dynamic motion and the numerous optimization variables involved. Traditional trajectory optimization methods address these challenges by formulating the problem as an optimization task, aiming to minimize cost functions, and to automatically discover contact sequences. Despite their structured approach, optimization-based methods face substantial difficulties, particularly because such formulations result in highly nonlinear and difficult to solve problems.
To address these limitations, we propose \acronym{}, a bi-level optimization method that combines traditional trajectory optimization with a black-box optimization scheme. \acronym{} at the higher level employs the Mixed Distribution Cross-Entropy Method to optimize both the gait sequence and the phase durations, thus simplifying the lower level trajectory optimization problem. This approach allows for fast solutions of complex gait optimization problems. Extensive evaluation in simulated environments demonstrates that \acronym{} can find solutions for biped, quadruped, and hexapod robots in under 10 seconds. This novel bi-level optimization scheme offers a promising direction for future research in automatic contact scheduling.
\end{abstract}

\section{Introduction}\label{sec:intro}
%
\begin{figure}[!htb]
    \centering
    \includegraphics[width=\linewidth]{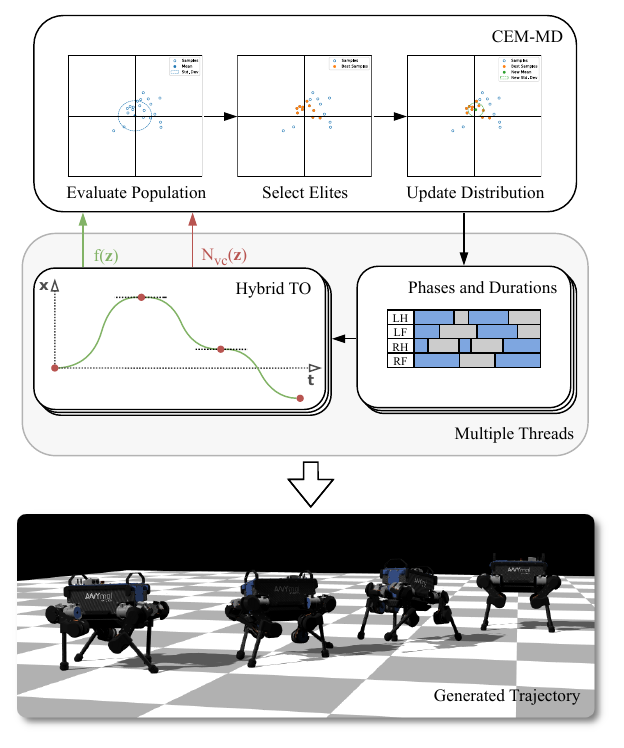}
    \vspace{-1em}
    \caption{Overview of \acronym{} method.}
    \label{fig:concept}
    \vspace{-2em}
\end{figure}
Legged robotic systems play a pivotal role in robotics due to their superior load-bearing capabilities, enhanced autonomy, and effective navigation on uneven terrain. These systems offer an optimal balance between autonomy and payload capacity, in contrast to flying and wheeled robots. Flying robots excel in diverse environments but face limitations in payload capacity and autonomy. On the other hand, wheeled robots are highly efficient in transporting heavy loads over long periods of time but are primarily restricted to even terrain. The biomimetic design of bipedal and humanoid robots further underscores their significance, as these systems are well-suited for environments typically designed for human mobility and interaction.

However, planning and executing the motion of legged systems has proven to be a challenge. Planning and optimizing the gaits for legged robots is inherently challenging due to the complexity of their dynamic motion and the numerous optimization variables involved~\cite{wensing2023optimization,carpentier2018multicontact}. Unlike wheeled robots, legged robots must maintain their balance while transitioning through various phases of movement, thus necessitating precise coordination of multiple joints and actuators. The need to adapt to diverse and unpredictable terrains further complicates gait optimization, as the robot must continually adjust its posture and force distribution to avoid instability and potential falls.

Trajectory optimization (TO) methods~\cite{wensing2023optimization} address the challenge of gait planning for legged robots by formulating it as an optimization problem. These methods aim to determine the optimal control inputs and joint trajectories that minimize a predefined cost function, such as energy consumption, stability, or smoothness of motion. Although computationally demanding, these methods provide a structured framework to achieve adaptive and efficient gait strategies, enabling legged robots to navigate complex environments~\cite{dai2014whole,winkler2018gait,posa2016optimization}.

Despite their success, applying optimization-based methods for selecting the type of gaits in legged robots presents substantial challenges. Contact-implicit methods~\cite{posa2014direct}, which incorporate contact dynamics directly into the optimization problem, lead to highly nonlinear and non-convex problems that are very hard to solve effectively. Additionally, methods based on mixed integer programming (MIP), which model gait decisions, such as gait transitions, with discrete variables, are time-consuming due to the combinatorial nature of the problem. The inclusion of integer variables exponentially increases the search space, making it difficult to find optimal solutions within a reasonable timescale, especially for high-dimensional systems, unless strong simplifications are made~\cite{fey20243d}. Thus, while optimization-based methods offer a powerful framework for gait selection, their practical application is hindered by computational demands and the complexity of accurately modeling dynamic interactions. Consequently, most successful optimization-based methods rely on a hybrid formulation of the dynamics, where we basically predefine the gait sequence by hand~\cite{winkler2018gait,mastalli2022agile}.

To overcome this limitation, we propose \methodname{}~(\acronym{}), a novel method that effectively combines traditional trajectory optimization with a black-box optimization method, enabling us to solve the gait optimization problem within seconds. \acronym{} formulates a bi-level optimization scheme. In particular, in the higher-level it employs the Mixed Distribution Cross-Entropy Method (CEM-MD)~\cite{xue2024logic}, which optimizes both discrete and continuous variables, to determine gait sequences and the duration of swing/stance phases. The key insight is that fixing the gait sequence and phases' duration simplifies the lower-level trajectory optimization problem significantly. Therefore, each individual in CEM-MD addresses a simpler trajectory optimization problem. Additionally, by leveraging the parallelization capabilities of CEM-MD, we can solve complex problems in a few seconds.

We extensively evaluate our novel methodology in simulated environments focusing on finding gait sequences for biped, quadruped and hexapod robots. The results showcase that \acronym{} is able to find solutions to difficult optimization problems in less than $10\,s$.

Overall, this original integration of black-box optimization and trajectory optimization paves the way for future research in automatic contact scheduling.
\section{Background}\label{sec:bg}
\subsection{Optimal Control and Trajectory Optimization}\label{sec:to}
The general optimal control problem can be defined as~\cite{wensing2023optimization}:
\begin{align}
    \argmin_{\boldsymbol{x}(t), \boldsymbol{u}(t)}&\mathcal{J}(\boldsymbol{x}(t), \boldsymbol{u}(t)) = \int_{t_0}^{t_f}\ell(\boldsymbol{x}(t), \boldsymbol{u}(t))dt + \ell_F(\boldsymbol{x}(t_f))\nonumber\\
    &\text{s.t.}\quad\quad\dot{\boldsymbol{x}}(t) = f(\boldsymbol{x}(t), \boldsymbol{u}(t))
\end{align}
where\footnote{We denote time derivatives with an upper dot: \emph{e.g.} $\dot{x}$.}
\begin{itemize}
    \item$\boldsymbol{x}(t)\in\mathbb{R}^{N_s}, \boldsymbol{u}(t)\in\mathbb{R}^{N_u}$ are the state and control trajectories
    \item$\mathcal{J}(\boldsymbol{x}(t), \boldsymbol{u}(t))$ is the ``cost function''
    \item$\ell(\boldsymbol{x}(t), \boldsymbol{u}(t))$ is the ``stage cost''
    \item$\ell_F(\boldsymbol{x}(t_f))$ is the ``terminal cost''
    \item$\dot{\boldsymbol{x}}(t) = f(\boldsymbol{x}(t), \boldsymbol{u}(t))$ are the ``dynamics constraints''
    \item we can potentially add more constraints
\end{itemize}
The above formulation results in an infinite-dimensional problem that is generally unsolvable on a computer. Numerous methods exist to transcribe this into a finite-dimensional problem, with discretization and direct collocation being the primary approaches~\cite{kelly2017introduction,wensing2023optimization}. Discretizing the above problem leads to the following generic formulation:
\begin{align}
    \argmin_{\boldsymbol{x}_{1:K}, \boldsymbol{u}_{1:K-1}}&\mathcal{J}(\boldsymbol{x}_{1:K}, \boldsymbol{u}_{1:K-1}) = \sum_{k=1}^{K-1}\ell(\boldsymbol{x}_k, \boldsymbol{u}_k) + \ell_F(\boldsymbol{x}_K)\nonumber\\
    &\text{s.t.}\quad\quad\boldsymbol{x}_{k+1} = f_{\text{discrete}}(\boldsymbol{x}_k, \boldsymbol{u}_k)
\end{align}
where $\boldsymbol{x}_k\in\mathbb{R}^{N_s}$ and $\boldsymbol{u}_k\in\mathbb{R}^{N_u}$. Direct collocation optimizes over knot points and links them using splines~\cite{kelly2017introduction}. Both methods are sufficiently versatile to yield strong solutions to the problem. For discretization, the problem is typically addressed using Differential Dynamic Programming~\cite{wensing2023optimization,mastalli2022feasibility,jallet2023proxddp}, whereas direct collocation employs general non-linear optimization solvers~\cite{kelly2017introduction}.
\subsection{Contact-Implicit Optimization}\label{sec:cio}
Contact-Implicit formulations include the contact forces as optimization variables, and form complementarity problems~\cite{posa2014direct}. A generic formulation can be described as follows:
\begin{align}
    \argmin_{\boldsymbol{x}_{1:K}, \boldsymbol{u}_{1:K-1}, \boldsymbol{\lambda}_{1:K-1}}&\sum_{k=1}^{K-1}\ell(\boldsymbol{x}_k, \boldsymbol{u}_k) + \ell_F(\boldsymbol{x}_K)\nonumber\\
    &\text{s.t.}\quad\quad\boldsymbol{x}_{k+1} = f_{\text{discrete}}(\boldsymbol{x}_k, \boldsymbol{u}_k, \boldsymbol{\lambda}_k)\nonumber\\
    &\quad\quad\quad\boldsymbol{\lambda}_k\geq 0\nonumber\\
    &\quad\quad\quad\phi(\boldsymbol{x}_k)\geq 0\nonumber\\
    &\quad\quad\quad\phi(\boldsymbol{x}_k)^T\boldsymbol{\lambda}_k=0
\end{align}
where
\begin{itemize}
    \item we have $N_m$ potential contact points
    \item$\boldsymbol{\lambda}_k\in\mathbb{R}^{N_m}$ the constraint forces acting along the surfaces' normal vectors
    \item$\phi(\boldsymbol{x}_k)$ are non-penetration constraints
\end{itemize}
Contact-Implicit methods have the ability to automatically discover contact switches and have produced extraordinary results~\cite{mordatch2012discovery,manchester2019contact,le2024fast,dai2014whole,posa2014direct}, but the optimization problems are hard to solve consistently and robustly, especially for longer time horizons (i.e. more than 100 ms).
%
%
\subsection{Single Rigid Body Dynamics Model}\label{sec:srbd}
\begin{figure}[!bt]
    \centering
    \includegraphics[width=0.9\linewidth]{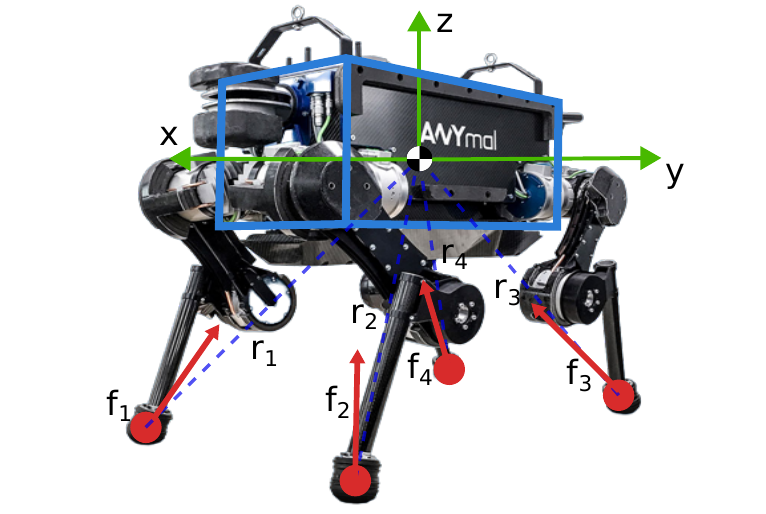}
    \caption{Single rigid body dynamics model with contacts. Image from~\cite{chatzilygeroudis2023evolving}.}
    \label{fig:srbd}
    \vspace{-1.75em}
\end{figure}

In this paper, we model walking robots using the Single Rigid Body Dynamics (SRBD) Model~\cite{winkler2018gait,wensing2023optimization,chatzilygeroudis2023evolving} (Fig.~\ref{fig:srbd}). In this model, we assume that the robot can be modeled as a single rigid body with constant moment of inertia and mass-less limbs/legs that can generate contact forces.
The rigid body has mass $m\in\mathbb{R}^+$ and moment of inertia $\boldsymbol{I}\in\mathbb{R}^{3\times 3}$. The body can be described by its linear position $\boldsymbol{p}_b\in\mathbb{R}^3$, linear velocity $\dot{\boldsymbol{p}}_b\in\mathbb{R}^3$, body orientation $\boldsymbol{R}\in SO(3)$ and angular velocity\footnote{We express the angular velocity in the body coordinate frame, while the rest of the variables are expressed in the world/inertial coordinate frame.} $\boldsymbol{\omega}\in\mathbb{R}^3$. Each leg $i$ has a linear position $\boldsymbol{p}_i\in\mathbb{R}^3$, and can generate contact forces $\boldsymbol{f}_i\in\mathbb{R}^3$. When leg $i$ is in a swing phase, it does not exert any contact force, $\boldsymbol{f}_i=\boldsymbol{0}$. Given all of the above, at each time step the following forces and torques are applied on the body:
\begin{align}
    &\boldsymbol{f}_{\text{total}} = \sum_i\boldsymbol{f}_i + m\,\boldsymbol{g},\nonumber\\[0.1cm]
    &\boldsymbol{\tau}_{\text{total}} = \sum_i\boldsymbol{r}_i\times\boldsymbol{f}_i,
\end{align}
where $\boldsymbol{g}$ is the gravity vector, $\boldsymbol{r}_i = \boldsymbol{p}_i - \boldsymbol{p}_b$, and $\times$ defines the cross product of two three-dimensional vectors. Given the forces and torques in the above equations, we can compute the linear and angular accelerations of the body as follows:
\begin{align}
    &\Ddot{\boldsymbol{p}} = \frac{\boldsymbol{f}_{\text{total}}}{m},\nonumber\\[0.15cm]
    &\dot{\boldsymbol{\omega}} = \boldsymbol{I}^{-1}(\boldsymbol{R}^{\top}\boldsymbol{\tau}_{\text{total}}-\boldsymbol{\omega}\times\boldsymbol{I}\boldsymbol{\omega}).
\end{align}
These equations are usually referred to as the \emph{Newton–Euler equations} for rigid body motion in three-dimensional space.
\vspace{-1em}
\section{Related Work}\label{sec:soa}
When trying to solve a trajectory optimization problem for a legged system, one has to decide how to solve the scheduling of the contact sequences. The easiest way, perhaps, of achieving this goal is to pre-specify a desired contact sequence~\cite{winkler2018gait,posa2016optimization,turski2023staged}: it should be noted that this does not necessarily mean restricting the timings and positions of contact points. In this case, the trajectory optimization problem becomes a \emph{hybrid trajectory optimization} problem instance. More specifically, the foot trajectory optimization problem is split into modes or phases with each phase being treated as a normal TO problem and connected to its neighboring phases via additional constraints. To make it clearer, let us consider an example with a monopod robot. The contact sequence, in this case, alternates between stance and swing phases for the single leg. During a swing phase, the foot cannot exert any force and thus a formulation where the robot is modeled without any external forces can be used. On the other hand, during a stance phase, the contact forces must be considered. We define additional constraints (at least in position and velocity) to ``link'' each phase with the next one. Despite its simplicity and obvious drawbacks (e.g. pre-defined contact sequence), this hybrid formulation is one of the most successful TO formulations for legged systems.

In~\cite{winkler2018gait}, the authors define a hybrid trajectory optimization problem and use a direct collocation transcription to solve the problem with a general non-linear programming solver. Their method is able to optimize for complex gaits in a few seconds, and the phases' duration can also be optimized over, which allows for a semi-automatic contact sequence discovery. In~\cite{turski2023staged}, the authors combine contact implicit optimization and hybrid TO in order automatically discover the contact sequence, while also keeping low computation times. Despite their success, hybrid TO methods do not allow for automatic contact sequences, and methods such as~\cite{turski2023staged} still require solving a contact implicit optimization problem which is difficult to do in a generic and robust fashion.

Contact implicit optimization methods are able to discover contact sequences automatically and have produced impressive results~\cite{mordatch2012discovery,manchester2019contact,le2024fast,dai2014whole,posa2014direct}. Although it is generally considered that contact implicit methods produce difficult to optimize problems, there have recently been serious advances in making them faster and more reliable~\cite{manchester2019contact,le2024fast,kim2023contact}. This is mostly achieved by intelligent splitting of the main problem (e.g. formulating the contact dynamics as a Linear Complementarity Problem~\cite{le2024fast} or relaxing the contact constraints~\cite{kim2023contact}) or using variational integrators~\cite{manchester2019contact}. Overall, contact implicit methods can discover effective gaits, but the underlying problems are still computationally demanding to solve, and/or require heavy engineering effort per task. \acronym{}, on the other hand, is able to automatically discover contact sequences in a few seconds on a standard CPU.

Another natural way of scheduling contact sequences is to define integer variables to model how many phases each leg can perform. This, obviously produces a Mixed Integer Programming (MIP) scheme and many methods have been proposed in this direction~\cite{deits2014footstep,valenzuela2016mixed,aceituno2017simultaneous,ding2020kinodynamic,fey20243d}. The main intuition of this approach is to approximate the non-convex mixed integer constraints with convex equivalents to then be able to use off-the-shelf fast MIP solvers. In~\cite{deits2014footstep} the authors incorporate kinematic reachability constraints to plan foot placements for the Atlas robot on uneven terrain. They restrict footstep positions to a convex region close to the robot. Valenzuela~\cite{valenzuela2016mixed} builds on this work by employing McCormick Envelopes~\cite{mccormick1976computability} to approximate bilinear terms in a centroidal dynamics model. Aceituno et al.~\cite{aceituno2017simultaneous} develop a MIP-based motion planning method for a quadruped robot, but restricts contact points to a region near the center of mass (COM). Finally, the authors in~\cite{fey20243d} propose a novel MIP formulation where they approximate each stance phase as impulsive and show that this simplification is enough to efficiently generate hopping behaviors for a quadruped robot. Overall, MIP methods can automatically discover contact sequences, but often have to make strong assumptions and simplifications in order to get to a tractable problem. On the contrary, \acronym{} is able to discover gaits in seconds without the need for extreme simplifications.

A few hybrid methods employing sampling-based planners~\cite{shkolnik2011bounding,manchester2011stable} have been proposed for solving the contact sequence and motion optimization problem for legged systems. These methods, however, depend on the modeling of specific cases by making many assumptions and simplifications, and are computationally demanding.

Overall, despite the efforts of the community, there is still no widely accepted method for automatic contact sequence and motion optimization for legged systems. \acronym{} aims at filling this gap by providing a versatile method that is able to automatic discover gait sequences along with optimized motions for any legged system in just a few seconds.
%
%
%
%
\section{\acronym{} Method}\label{sec:method}
\subsection{Overview}\label{sec:overview}
\acronym{} effectively combines black-box optimization with trajectory optimization by establishing a bi-level optimization framework (Fig.~\ref{fig:concept}). The high-level optimizer, CEM-MD (Sec.~\ref{sec:cemmd}), determines the number of stance and swing phases for each leg as well as the duration of each phase. The low-level optimizer addresses a trajectory optimization problem where the leg phases and durations are fixed. This approach allows CEM-MD to manage the combinatorial complexity of phase selection, while the low-level solver tackles the more manageable hybrid trajectory optimization problems.
\subsection{Mixed Distribution Cross Entropy Method}\label{sec:cemmd}
We take inspiration from recent work and utilize a Mixed Distribution Cross Entropy Method~\cite{xue2024logic} to solve the mixed integer optimization problem of selecting the number of phases and their duration. CEM-MD follows the same principles as the original CEM algorithm, and employs a categorical distribution, alongside the continuous distribution, for modeling the discrete variables~(Algo.~\ref{algo:cemmd}).

In the original CEM algorithm, the population is modeled via a Gaussian distribution, whereas in CEM-MD we model the population via two independent distributions: one Gaussian for the continuous variables and a categorical one for the discrete variables.
The algorithm starts each iteration by sampling from the distributions a population of $N$ candidate solutions (lines~\ref{algo:sample}-\ref{algo:pop} in Algo.~\ref{algo:cemmd}). It then evaluates the candidate solutions (line~\ref{algo:eval} in Algo.~\ref{algo:cemmd}). It then keeps the $M$ best candidate solutions and uses them to update the distributions. Finally, CEM-MD performs several iterations until some convergence criteria is met.
\begin{algorithm}
    \caption{Mixed Distribution Cross Entropy Method (CEM-MD)}\label{algo:cemmd}
    \begin{algorithmic}[1]
        \Procedure{CEM-MD}{$N, K, M, J, C$}
            \State Initialize $\boldsymbol{\mu}_1, \boldsymbol{\sigma}_1, \boldsymbol{p}_1$
            \For {$k=1\to K$}\Comment{$K$ iterations}
                \State $\mathcal{P}_{\text{cont}} = \{\boldsymbol{z}^c_1, \dots, \boldsymbol{z}^c_N\}$, where $\boldsymbol{z}^c_i\sim\mathcal{N}(\boldsymbol{\mu}_k, \text{diag}(\boldsymbol{\sigma}_k^2))$\Comment{Generate continuous population of size $N$}\label{algo:sample}
                \State $\mathcal{P}_{\text{discrete}} = \{\boldsymbol{z}^d_1, \dots, \boldsymbol{z}^d_N\}$, where $\boldsymbol{z}^d_i\sim\text{Categorical}(\boldsymbol{p}_k)$\Comment{Generate discrete population of size $N$}\label{algo:sample_d}
                \State$\mathcal{P} = \{\boldsymbol{z}_1 = (\boldsymbol{z}^c_1, \boldsymbol{z}^d_1), \dots, \boldsymbol{z}_N = (\boldsymbol{z}^c_N, \boldsymbol{z}^d_N)\}$\label{algo:pop}
                \State$\mathcal{F} = \{J(\boldsymbol{z}_1), \dots, J(\boldsymbol{z}_N)\}$\Comment{Evaluate individuals with objective function $J$}\label{algo:eval}
                \State$\mathcal{P}_{\text{elite}} = \{\boldsymbol{z}^{\text{elite}}_1, \dots, \boldsymbol{z}_M^{\text{elite}}\}$\Comment{Select $M$ elites}
                \State$\boldsymbol{\mu}_{k+1} = \frac{1}{M}\sum_{\boldsymbol{z}^{\text{elite},c}_i\in\mathcal{P}_{\text{elite}}}\boldsymbol{z}^{\text{elite},c}_i$\Comment{Update continuous mean}
                \State$\boldsymbol{\sigma}^2_{k+1} = \frac{1}{M}\sum_{\boldsymbol{z}^{\text{elite},c}_i\in\mathcal{P}_{\text{elite}}}(\boldsymbol{z}^{\text{elite},c}_i-\boldsymbol{\mu}_k)^2$\Comment{Update continuous variance}
                \For {$i=1\to C$}\Comment{$C$ categories}
                    \State$\boldsymbol{p}_{k+1}^i = \frac{\text{count\_occurrences}(\text{Category}_i)}{M}$
                \EndFor
            \EndFor
        \EndProcedure
    \end{algorithmic}
\end{algorithm}
%
\subsubsection{Parallel Computation}
CEM-MD, like the original CEM algorithm, can take advantage of multi-core computers and compute in parallel the function evaluations (in particular, line~\ref{algo:eval} in Algo.~\ref{algo:cemmd}). This parallelization can significantly accelerate the wall-time of the process by enabling simultaneous evaluations of numerous candidate solutions. This scalability and speedup can be particularly beneficial in realistic applications, where fast responses are crucial, such as in our problem: gait optimization for legged systems.

Moreover, provided that we can run the trajectory optimization process on the GPU (e.g. see~\cite{adabag2023mpcgpu,jeon2024cusadi}), we can foresee an implementation of \acronym{} completely on the GPU allowing us to increase the parallelization and speedup of the whole pipeline. In this case, we can also considerably increase the population size of CEM-MD and get more optimized results. We leave this endeavor for future work.

\subsubsection{Optimizing Gaits}
In order for CEM-MD to produce feasible gaits, we need to make sure that the underlying TO problems are being solved. For this reason, we can use the minimization of the number of violated constraints of the TO problem as an objective function for CEM-MD. In most experiments, we exit the optimization process when CEM-MD finds a solution with zero violated constraints; this means that CEM-MD found a TO initialization that produces a feasible result. We can also choose different objectives for CEM-MD, for example: a) minimizing the number of phases, b) minimizing the feet forces. In all cases, we need to ensure that we have zero violated constraints. For this reason, we use the following generic objective function (we assume that CEM-MD maximizes the objective function):
\begin{align}\label{eq:cemmd_obj}
    J(\boldsymbol{z}) = f(\boldsymbol{z}) - \alpha_{\text{penalty}}N_{\text{vc}}(\boldsymbol{z})
\end{align}%
where $N_{\text{vc}}(\boldsymbol{z})$ is the number of violated constraints of candidate solution $\boldsymbol{z}$, $\alpha_{\text{penalty}}$ is a penalty factor to ensure that CEM-MD prioritizes finding feasible solutions, and $f(\boldsymbol{z})$ is any other objective function that we might care about.
%
\subsection{Trajectory Optimization with SRBD}\label{sec:trajopt}
Each candidate solution of CEM-MD solves a trajectory optimization problem with fixed number of phases and durations (i.e. when calling $J(\cdot)$). We follow the formulation of Winkler~\emph{et al.}~\cite{winkler2018gait} and:
\begin{itemize}
    \item[a)] Model the robot as a single rigid body mass with legs of negligible mass. (Sec.~\ref{sec:srbd});
    \item[b)] Adopt Winkler \emph{et al.}~\cite{winkler2018gait} phase-based formulation for contact switching (the phases are coming from CEM-MD, and are thus fixed);
    \item[c)] Parameterize the body pose, foot positions and foot forces with multiple cubic Hermite polynomials.
\end{itemize}
Overall, we have the following optimization problem (omitting the cubic polynomials for clarity):
{\small
\begin{align*}\label{eq:TOWR}
    \text{find}\quad&\boldsymbol{r}(t), \,\,\boldsymbol{r}:\mathbb{R}\to\mathbb{R}^3,\hfill&\text{(Body positions)}\nonumber\\
    &\boldsymbol{\theta}(t), \,\, {\boldsymbol{\theta}}:\mathbb{R}\to\mathbb{R}^3,\hfill&\text{(Body Euler angles)}\nonumber\\
    &\boldsymbol{p}(t), \,\, \boldsymbol{p} :\mathbb{R}\to\mathbb{R}^3,\hfill&\text{(Foot position)}\nonumber\\
    &\boldsymbol{f}(t), \,\, \boldsymbol{f}:\mathbb{R}\to\mathbb{R}^3,\hfill&\text{(Foot force)}\nonumber\\
    \text{s.t.}\quad&\text{srbd}(\boldsymbol{r},\boldsymbol{\theta},\boldsymbol{p},\boldsymbol{f}) = \{\ddot{\boldsymbol{r}}, \ddot{\boldsymbol{\theta}}\},\hfill&\text{(Dynamics)}\nonumber\\
    &\{\boldsymbol{r}(0),\boldsymbol{\theta}(0)\} = \{\boldsymbol{r}_{\text{init}},\boldsymbol{\theta}_{\text{init}}\},\hfill&\text{(Initial State)}\nonumber\\
    &\{\boldsymbol{r}(T),\boldsymbol{\theta}(T)\} = \{\boldsymbol{r}_{\text{goal}},\boldsymbol{\theta}_{\text{goal}}\},\hfill&\text{(Goal State)}\nonumber\\
    &\boldsymbol{p}(t)\in\mathcal{B}\bigl(\boldsymbol{r}(t),\boldsymbol{\theta}(t)\bigr),\hfill&\text{(Bounds wrt body)}\nonumber\\
    &\boldsymbol{\dot{p}}(t)=\boldsymbol{0},\text{ for  } t\in\text{Contact},\hfill&\text{(No slip)}\nonumber\\
    &\boldsymbol{p}(t)\in\mathcal{T},\text{ for  } t\in\text{Contact},\hfill&\text{(Contact on terrain)}\nonumber\\
    &\boldsymbol{f}(t)\in\mathcal{F},\text{ for  } t\in\text{Contact},\hfill&\text{(Pushing force/friction cone)}\nonumber\\
    &\boldsymbol{f}(t)=\boldsymbol{0},\text{ for  } t\notin\text{Contact},\hfill&\text{(No force in air)}
\end{align*}
}%
We use the Ipopt solver~\cite{wachter2006implementation} to optimize the above problem. It is important to note that even though in this paper we use the SRBD formulation and Winkler \emph{et al.}'s transcription, \acronym{} is not limited to those and can be utilized with any other transcription and/or model.
\vspace{-0.5em}
\subsection{Heuristic Sampling Bias}\label{sec:bias_sampling}
Although our CEM-MD formulation efficiently finds gait sequences within a few iterations, it encounters difficulties when the number of potential combinations is exceedingly high. To address this, we implement a straightforward yet effective heuristic to reduce the possible combinations. The key idea is that \emph{having significantly different numbers of phases per leg generally leads to ill-defined trajectory optimization problems that cannot be solved}. Therefore, rather than considering all possible leg phases combinations, we restrict our selection to some of those where the difference in the number of phases between the legs is less than three.

This considerably reduces the number of possible combinations, while keeping useful combinations that can yield good results. For example, let us suppose we have a quadruped robot and want the solver to choose among 7 phases for each leg. We have $7^4 = 2401$ total combinations. In our heuristic, we choose the following triplets: all legs are allowed to have at the same time a) 1, 2 or 3 phases, b) 3, 4 or 5 phases, and c) 5, 6 or 7 phases. Thus we now have $3\times 3^4 = 243$ total combinations. Even though we have removed many possible combinations, the removed combinations are the ones most likely to produce ill-conditioned TO problems (e.g. one leg having 1 phase, another having 7 phases, etc.). At the same time, the remaining combinations are enough to allow for adequate exploration of the gait space by \acronym{}. If not specified otherwise, when we refer to \acronym{}, we assume that CEM-MD with the heuristic sampling is used.
%
%
\vspace{-0.5em}
\section{Experiments and Results}\label{sec:exps}
In order to validate \acronym{}, we devise several simulated experiments. More precisely, we aim at answering the following questions:
\begin{enumerate}
    \item Is \acronym{} able to find gait sequences for several scenarios and how does it compare to Contact-Implicit and hybrid methods?
    \item Why is \acronym{} better than computing in parallel all possible gait sequence scenarios and taking the best?
    \item Can \acronym{} find optimal gait sequences and durations?
\end{enumerate}

Overall, the experiments showcase the versatility of \acronym{} in finding effective gaits for different robots within just a few seconds. Videos of the generated trajectories played forward (i.e visualized, not simulated) are available at {\url{https://nosalro.github.io/cregopt}}.
\subsection{Practical Considerations}\label{sec:practical}
From a practical point of view, we are making the following choices:
\begin{enumerate}
    \item We model the number of phases per leg as independent categorical variables. This choice was made since the implementation and engineering effort was smaller. Moreover, initial experiments with one big variable (i.e., allowing for correlations) yielded similar results.
    \item Since we cannot change the number of optimization variables on the fly, and the duration vector size depends on the number of phases per leg, we create a duration vector with the maximum possible size and specific indexing for each leg and phase.
    \item Since each duration should always be positive, we are optimizing them in log space, and we are taking the exponential before passing them to the TO problem.
\end{enumerate}
All experiments are performed on an Intel i7-10700K 16-core CPU @ 3.80GHz. The hyper-parameters used for CEM-MD in our experiments are shown in Table~\ref{tab:hyperparams}.

\begin{table}[!tbh]
    \vspace{-1em}
    \caption{Hyperparameters for CEM-MD}
    \vspace{-1em}
    \label{tab:hyperparams}
    \centering
    \resizebox{\linewidth}{!}{%
    \begin{tabular}{lccc}
    \toprule
    &\multicolumn{3}{c}{Hyperparameters per Experiment}\\
    \cmidrule(){2-4}
    Parameter     &  Sec.~\ref{sec:simple_exps} & Sec.~\ref{sec:scaling} & Sec.~\ref{sec:go_exps} \\
    \midrule
    Population size ($N$) &  16 & 16 & 64 \\
    Elites size ($M$)    &  12 & 12 & 32  \\
    Maximum Iterations ($K$) & 20 & 20 & 50 \\
    Termination Condition & \multicolumn{2}{c}{\emph{Feasible solution}} & \emph{Convergence}\\
    $\boldsymbol{\mu}_1$  &  -1.75 ($\approx 0.17\,s$) & -1.75 & -1.75 \\
    Minimum value & -2.5 ($\approx 0.1\,s$) & -2.5 & -2.5 \\
    Maximum value & -1 ($\approx 0.36\,s$) & -1 & -1 \\
    $\boldsymbol{\sigma}_1$  &  1 & 1 & 1 \\
    $\boldsymbol{p}_1$ & $\frac{1}{C}$ & $\frac{1}{C}$ & $\frac{1}{C}$ \\
    Minimum Stance Phases & 5 & 5 & 3 \\
    Maximum Stance Phases & 13 & 13 & 11 \\
    $\alpha_{\text{penalty}}$ & 100 & 100 & 1000 \\
    \bottomrule
    \end{tabular}
    }
    \vspace{-1em}
\end{table}
\subsection{Automatic Gait Sequence Experiments (Question 1)}\label{sec:simple_exps}
In order to answer the first question, we devise three different scenarios: a) a quadruped robot\footnote{We are using TOWR's~\cite{winkler2018gait} SRBD model of the ANYmal robot.} has to go forward 6 meters on a flat terrain, b) a quadruped robot has to go forward 6 meters while rotating around the z-axis on a flat terrain, and c) a quadruped robot has to go forward 6 meters while rotating around the z-axis and climbing a step. We compare our method to a state-of-the-art hybrid TO method (with time durations optimization), TOWR~\cite{winkler2018gait}, and a contact-implicit formulation.


In each scenario, we define 20 targets that are all 6 meters away from the starting point and reside on an arc (angle from $-\frac{\pi}{4}$ to $\frac{\pi}{4}$), so that we can define statistics for all methods and provide meaningful results on the success rate of each one.
We allow 40 seconds of wall-time for Ipopt to find a solution during the TO step and consider as successful the cases when the optimizer managed to find an optimal solution within the time budget.

\begin{table}[!tbh]
    \caption{Success Rates for Experiments of Sec.~\ref{sec:simple_exps}}
    \label{tab:sc_simple}
    \centering
    \begin{tabular}{lrrr}
    \toprule
    &\multicolumn{3}{c}{Success Rate}\\
    \cmidrule(){2-4}
    Algorithm     &  Scenario 1 & Scenario 2 & Scenario 3 \\
    \midrule
    \textbf{\acronym{}} &  \textbf{100\%} & \textbf{100\%} & \textbf{100\%} \\
    TOWR    &  95\% & 95\% & 95\%  \\
    Contact Implicit & 0\% & 0\% & 0\% \\
    \bottomrule
    \end{tabular}
    \vspace{-1em}
\end{table}

\begin{figure}[!bt]
    \centering
    \includegraphics[width=\linewidth]{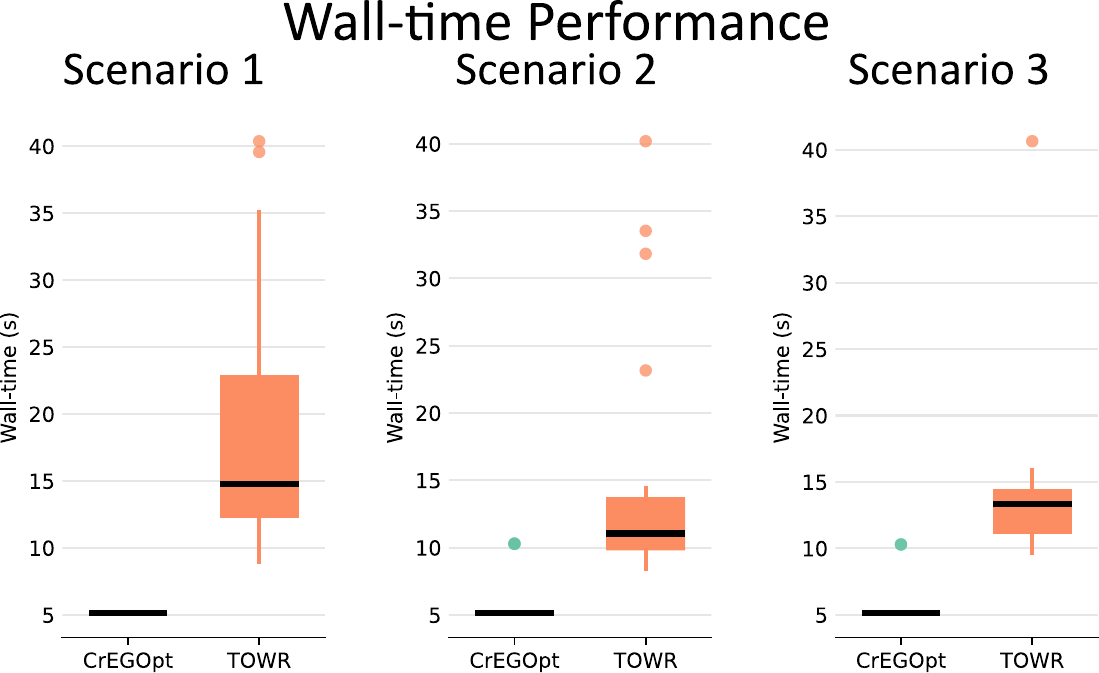}
    \vspace{-1.75em}
    \caption{Wall-time performance for experiments of Sec.~\ref{sec:simple_exps}. Contact-Implicit Optimization always maxed out the maximum wall-time. The box plots show the median (black line) and the interquartile range (25th and 75th percentiles) over 20 replicates; the whiskers extend to the most extreme data points not considered outliers, and outliers are plotted individually.}
    \label{fig:wall_simple}
    \vspace{-1.5em}
\end{figure}

The results showcase that across all scenarios, \acronym{} is able to always discover effective gait sequences, while TOWR fails around 5\% of the time and contact implicit optimization is not able to find a feasible solution within the maximum time budget we allow (Tab.~\ref{tab:sc_simple}). Additionally, \acronym{} outperforms both TOWR and contact implicit optimization in wall-time performance (Fig.~\ref{fig:wall_simple}) and is able to find gait sequences in less than 10 seconds. Interestingly, \acronym{} with our heuristic sampling was able to find a feasible solution in just 1 or 2 CEM-MD iterations.
Overall, the results of this section showcase that \acronym{} is able to effectively discover gait sequences in just a few seconds.

\subsection{Scaling Experiments (Question 2)}\label{sec:scaling}
%
In this section, we aim at answering the second question. In essence, since we are randomly sampling candidate solutions for CEM-MD, the following question naturally comes up: \emph{Why not run in parallel all possible candidate gait sequences and select the best one?} There are two main reasons why using \acronym{} is better than enumerating all possible gait sequences.

\begin{figure}[!bt]
    \centering
    \includegraphics[width=\linewidth]{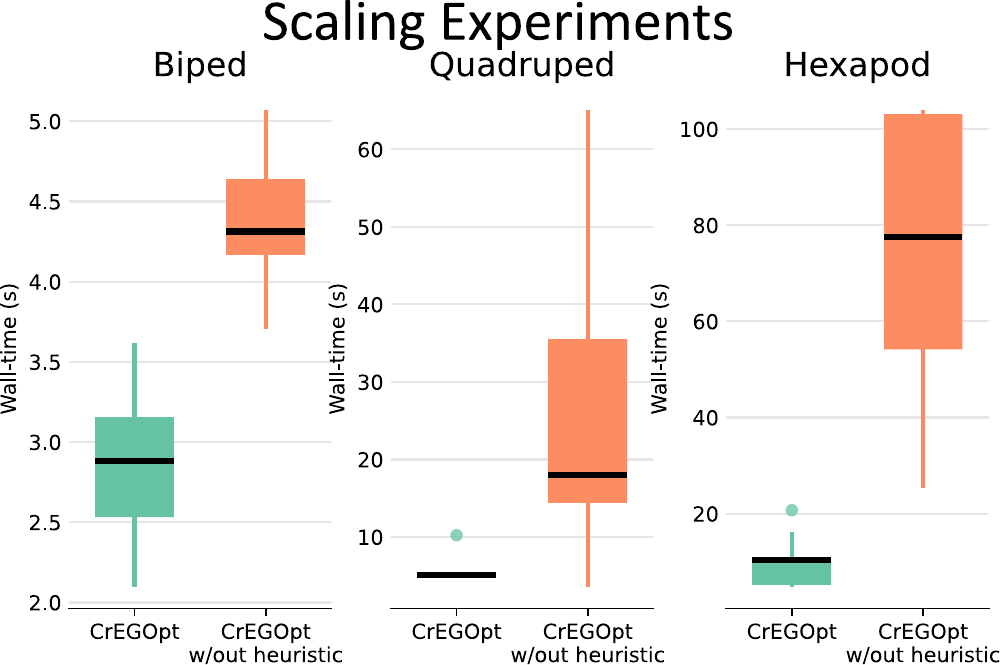}
    \vspace{-1.75em}
    \caption{Scaling experiments (Sec.~\ref{sec:scaling}, 20 replicates). \acronym{} is able to keep similar wall-times performances ($\approx$7\,s) even when increasing the number of legs. On the contrary, \acronym{} without the proposed heuristic sampling fails to find effective gaits as the number of legs increases.}
    \label{fig:wall_scaling}
    \vspace{-1.75em}
\end{figure}
\begin{figure}[!htb]
    \centering
    \includegraphics[width=\linewidth]{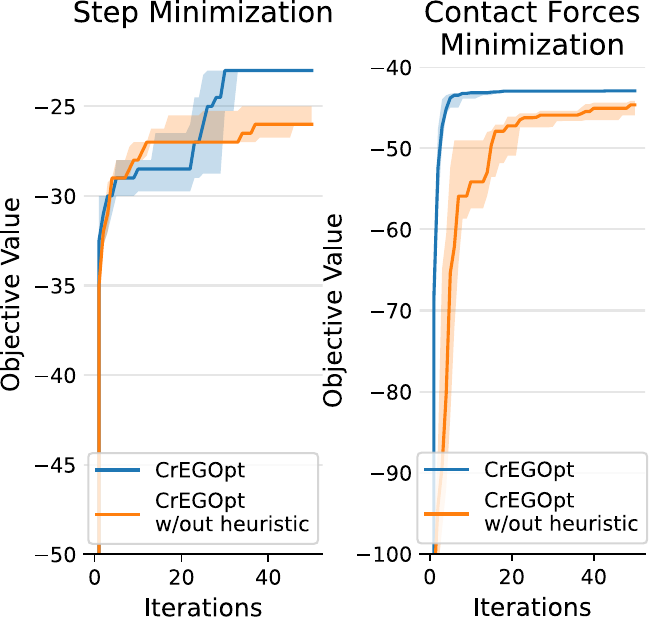}
    \vspace{-1.75em}
    \caption{\acronym{} optimization results (Sec.~\ref{sec:go_exps}). Solid lines are the median over 10 replicates and the shaded regions are the regions between the 25-th and 75-th percentiles.}
    \label{fig:opt_results}
\end{figure}
\begin{figure*}[!htb]
    \centering
    \includegraphics[width=\linewidth]{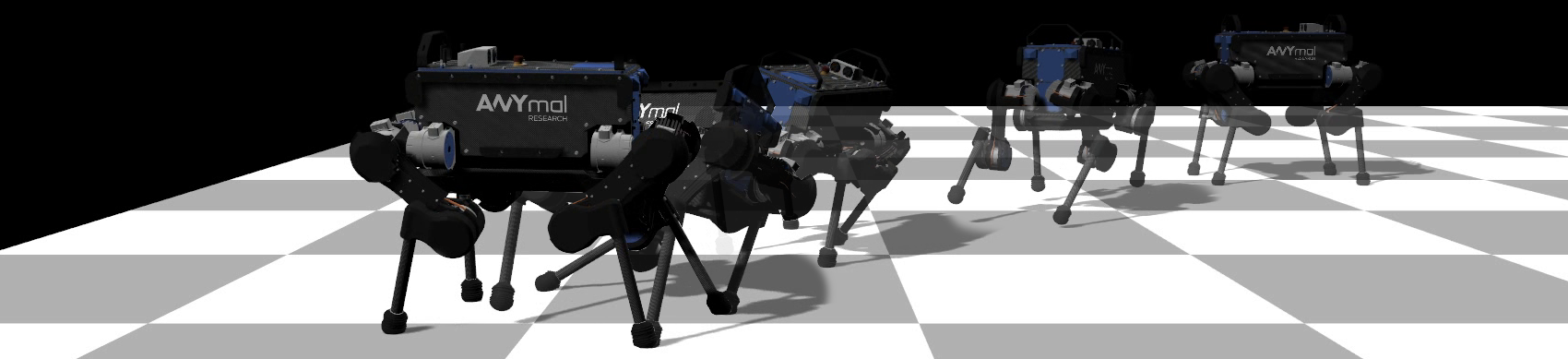}
    \vspace{-1.5em}
    \caption{Typical behavior returned from \acronym{} when minimizing the number of steps (Sec.~\ref{sec:go_exps}).}
    \label{fig:min_steps}
    \vspace{-1.5em}
\end{figure*}

\subsubsection{Scaling}
\acronym{} scales much better with the number of legs. This is easy to see with an example. Let us assume that we would like to give a maximum number of 7 phases per leg. For a biped robot, we would have $7^2 = 49$ possible combinations, $7^4 = 2401$ total combinations for a quadruped, and $7^6 = 117649$ possible combinations for a hexapod. While evaluating all possible combinations in parallel might be viable for the biped, it would be prohibitive to do so for a hexapod robot. Moreover, \acronym{} in combination with our heuristic sampling usually converges in just a few CEM-MD iterations, meaning that we only need to evaluate a few dozen combinations. To experimentally validate this,
we devise a set of experiments with a biped\footnote{We are using TOWR's~\cite{winkler2018gait} SRBD model of the biped HyQ robot.}, a quadruped and a hexapod robot\footnote{We are using an SRBD model of a real servo-based hexapod robot~\cite{chatzilygeroudis2018using}.} and compare the wall-time performance.

\begin{table}[!tbh]
    \vspace{-1em}
    \caption{Success Rates for Experiments of Sec.~\ref{sec:scaling}}
    \label{tab:sc_scaling}
    \centering
    \begin{tabular}{lrrr}
    \toprule
    &\multicolumn{3}{c}{Success Rate}\\
    \cmidrule(){2-4}
    Algorithm     &  Biped & Quadruped & Hexapod \\
    \midrule
    \textbf{\acronym{}} &  \textbf{100\%} & \textbf{100\%} & \textbf{100\%} \\
    \acronym{} w/out heuristic    &   \textbf{100\%} & 55\% & 50\%  \\
    \bottomrule
    \end{tabular}
\end{table}

The results showcase that \acronym{} is able to find feasible gait sequences for all robots in about the same wall-time (Fig.~\ref{fig:wall_scaling}). Moreover, when not using our heuristic sampling (Sec.~\ref{sec:bias_sampling}), CEM-MD has significantly lower success rate in the quadruped and hexapod robot cases (Tab.~\ref{tab:sc_scaling}). Overall, \acronym{} is able to automatically discover walking gaits for all robots in under 10\,s.
\subsubsection{Duration Optimization}
\acronym{} does not only find the gait sequence, but also optimizes for the duration of each phase. Since time durations are continuous variables, the use of an optimization algorithm, such as CEM-MD, allows \acronym{} to discover gait patterns. In essence, \acronym{} provides a versatile framework for gait sequence and motion optimization for legged systems.
\vspace{-0.5em}
\subsection{Gait Optimization Experiments (Question 3)}\label{sec:go_exps}
In this section, we aim at answering the 3rd question. For this reason, we devise a scenario where a quadruped robot needs to reach a target ($\approx$~4 meters away) while rotating around the z-axis. Here we are interested in maximizing a specific objective function while ensuring feasible TO problems. We are going to use Eq.~\eqref{eq:cemmd_obj} and define a few different $f(\cdot)$ functions. Moreover, we will increase the population size of CEM-MD and let it run for more iterations. In this manner, we will be able to observe if \acronym{} is able to optimize specific objective functions.
\subsubsection{Minimizing the Number of Steps} We first define an objective function that attempts to minimize the total number of steps for all legs. In other words, $f(\boldsymbol{z}) = -\sum_{i=1}^Cz^c_i$, where $z^c_i$ is the $i$-th component of $\boldsymbol{z}^c$, and $\boldsymbol{z} = (\boldsymbol{z}^c, \boldsymbol{z}^d)$. Fig.~\ref{fig:opt_results} showcases that \acronym{} is able to optimize the objective function and refine it until it converges to a (possibly local) minimum. Moreover, \acronym{} converges much faster than simple CEM-MD. Fig.~\ref{fig:min_steps} shows a typical resulting behavior. We see that \acronym{} is able to discover a fast ``galloping''-like gait in order to minimize the number of steps.

\subsubsection{Minimizing the Contact Forces} 
\begin{figure*}[!htb]
    \centering
    \includegraphics[trim={0 1cm 0 1cm},clip,width=\linewidth]{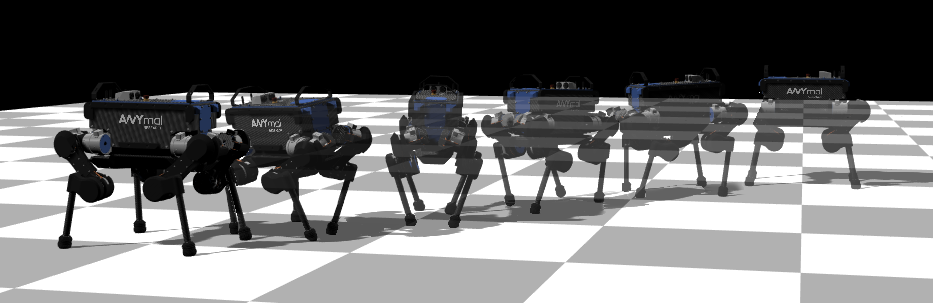}
    \vspace{-1.5em}
    \caption{Typical behavior returned from \acronym{} when minimizing the contact forces (Sec.~\ref{sec:go_exps}).}
    \label{fig:min_force}
    \vspace{-1.5em}
\end{figure*}

We now define an objective function that attempts to minimize the ground reaction forces for all legs. We define $f(\boldsymbol{z})$ as the negative sum of all contact forces along the ground normal direction; in essence, we get the result from the TO problem at candidate $\boldsymbol{z}$, sample $N_{\text{sample}}$ points along the force trajectory, and return the negative sum of the forces acting along the normal direction. Fig.~\ref{fig:opt_results} showcases that \acronym{} is able to optimize this new objective function as well and it converges to a local minimum. Fig.~\ref{fig:min_force} shows a typical behavior produced with this objective function. We see that \acronym{} is able to produce a completely different gait (compared to the previous one) in order to minimize this new objective function.

%
%
\vspace{-0.5em}
\section{Conclusion}\label{sec:conclusion}
%
\acronym{} effectively combines black-box optimization and trajectory optimization in order to optimize gait sequences and motions for legged systems. Overall, this novel method is versatile and can find effective gait sequences for multiple legged systems in just a few seconds.

Although effective, \acronym{} has two main limitations. First, we cannot get any convergence guarantees or even convergence analysis of CEM-MD. Thus, there might be cases where it fails completely to find solutions. Secondly, our current implementation of \acronym{} cannot be used for real-time sensitive applications. Using the simplified SRBD model and maximum parallelization on our CPU, we weren't able to find solutions faster than 2\,s (i.e. 0.5\,Hz). The main bottleneck is the optimization time taken by failed TO problems in the lower-level. We aim at investigating this in future work; for example, we can devise criteria to stop earlier or predict failing TO problems before they exceed the maximum iterations/wall-time.

Moreover, we plan to exploit methods that can perform the trajectory optimization process on the GPU (e.g.~\cite{adabag2023mpcgpu,jeon2024cusadi}), and implement \acronym{} entirely on the GPU, thereby enhancing the parallelization and speed of the entire pipeline. This would also allow for a significant increase in the population size of CEM-MD, leading to more optimized results.

Overall, our novel combination of black-box optimization and trajectory optimization sets the stage for future research in automated contact scheduling. Thus, we also foresee use cases of our approach in non-legged systems; e.g. using CEM-MD and trajectory optimization for automatic contact scheduling in manipulation scenarios.





\vspace{-0.5em}
\section*{Acknowledgment}
This work was supported by the Hellenic Foundation for Research and Innovation (H.F.R.I.) under the ``3rd Call for H.F.R.I. Research Projects to support Post-Doctoral Researchers'' (Project Acronym: NOSALRO, Project Number: 7541). The authors would like to thank Dionis Totsila for helping with the creation of the figures.
%
\vspace{-0.5em}
\bibliographystyle{ieeetr}
\bibliography{references}

\begin{thebibliography}{10}

\bibitem{wensing2023optimization}
P.~M. Wensing, M.~Posa, Y.~Hu, A.~Escande, N.~Mansard, and A.~Del~Prete,
  ``Optimization-based control for dynamic legged robots,'' {\em IEEE
  Transactions on Robotics}, 2023.

\bibitem{carpentier2018multicontact}
J.~Carpentier and N.~Mansard, ``Multicontact locomotion of legged robots,''
  {\em IEEE Transactions on Robotics}, vol.~34, no.~6, pp.~1441--1460, 2018.

\bibitem{dai2014whole}
H.~Dai, A.~Valenzuela, and R.~Tedrake, ``Whole-body motion planning with
  centroidal dynamics and full kinematics,'' in {\em IEEE-RAS International
  Conference on Humanoid Robots}, 2014.

\bibitem{winkler2018gait}
A.~W. Winkler, C.~D. Bellicoso, M.~Hutter, and J.~Buchli, ``Gait and trajectory
  optimization for legged systems through phase-based end-effector
  parameterization,'' {\em IEEE Robotics and Automation Letters}, vol.~3,
  no.~3, pp.~1560--1567, 2018.

\bibitem{posa2016optimization}
M.~Posa, S.~Kuindersma, and R.~Tedrake, ``Optimization and stabilization of
  trajectories for constrained dynamical systems,'' in {\em IEEE International
  Conference on Robotics and Automation (ICRA)}, 2016.

\bibitem{posa2014direct}
M.~Posa, C.~Cantu, and R.~Tedrake, ``A direct method for trajectory
  optimization of rigid bodies through contact,'' {\em The International
  Journal of Robotics Research}, vol.~33, no.~1, pp.~69--81, 2014.

\bibitem{fey20243d}
N.~Fey, R.~J. Frei, and P.~M. Wensing, ``3d hopping in discontinuous terrain
  using impulse planning with mixed-integer strategies,'' {\em IEEE Robotics
  and Automation Letters}, 2024.

\bibitem{mastalli2022agile}
C.~Mastalli, W.~Merkt, G.~Xin, J.~Shim, M.~Mistry, I.~Havoutis, and
  S.~Vijayakumar, ``Agile maneuvers in legged robots: a predictive control
  approach,'' {\em arXiv preprint arXiv:2203.07554}, 2022.

\bibitem{xue2024logic}
T.~Xue, A.~Razmjoo, S.~Shetty, and S.~Calinon, ``Logic-skill programming: An
  optimization-based approach to sequential skill planning,'' {\em arXiv
  preprint arXiv:2405.04082}, 2024.

\bibitem{kelly2017introduction}
M.~Kelly, ``An introduction to trajectory optimization: How to do your own
  direct collocation,'' {\em SIAM Review}, 2017.

\bibitem{mastalli2022feasibility}
C.~Mastalli, W.~Merkt, J.~Marti-Saumell, H.~Ferrolho, J.~Sol{\`a}, N.~Mansard,
  and S.~Vijayakumar, ``A feasibility-driven approach to control-limited ddp,''
  {\em Autonomous Robots}, vol.~46, no.~8, pp.~985--1005, 2022.

\bibitem{jallet2023proxddp}
W.~Jallet, A.~Bambade, E.~Arlaud, S.~El-Kazdadi, N.~Mansard, and J.~Carpentier,
  ``Proxddp: Proximal constrained trajectory optimization,'' {\em Preprint},
  2023.

\bibitem{mordatch2012discovery}
I.~Mordatch, E.~Todorov, and Z.~Popovi{\'c}, ``Discovery of complex behaviors
  through contact-invariant optimization,'' {\em ACM Transactions on Graphics
  (ToG)}, vol.~31, no.~4, pp.~1--8, 2012.

\bibitem{manchester2019contact}
Z.~Manchester, N.~Doshi, R.~J. Wood, and S.~Kuindersma, ``Contact-implicit
  trajectory optimization using variational integrators,'' {\em The
  International Journal of Robotics Research}, vol.~38, no.~12-13,
  pp.~1463--1476, 2019.

\bibitem{le2024fast}
S.~Le~Cleac'h, T.~A. Howell, S.~Yang, C.-Y. Lee, J.~Zhang, A.~Bishop,
  M.~Schwager, and Z.~Manchester, ``Fast contact-implicit model predictive
  control,'' {\em IEEE Transactions on Robotics}, 2024.

\bibitem{chatzilygeroudis2023evolving}
K.~I. Chatzilygeroudis, C.~G. Tsakonas, and M.~N. Vrahatis, ``Evolving dynamic
  locomotion policies in minutes,'' in {\em International Conference on
  Information, Intelligence, Systems \& Applications (IISA)}, 2023.

\bibitem{turski2023staged}
M.~R. Turski, J.~Norby, and A.~M. Johnson, ``Staged contact optimization:
  Combining contact-implicit and multi-phase hybrid trajectory optimization,''
  in {\em IEEE/RSJ International Conference on Intelligent Robots and Systems
  (IROS)}, 2023.

\bibitem{kim2023contact}
G.~Kim, D.~Kang, J.-H. Kim, S.~Hong, and H.-W. Park, ``Contact-implicit mpc:
  Controlling diverse quadruped motions without pre-planned contact modes or
  trajectories,'' {\em arXiv preprint arXiv:2312.08961}, 2023.

\bibitem{deits2014footstep}
R.~Deits and R.~Tedrake, ``Footstep planning on uneven terrain with
  mixed-integer convex optimization,'' in {\em IEEE-RAS international
  conference on humanoid robots}, 2014.

\bibitem{valenzuela2016mixed}
A.~K. Valenzuela, {\em Mixed-integer convex optimization for planning
  aggressive motions of legged robots over rough terrain}.
\newblock PhD thesis, Massachusetts Institute of Technology, 2016.

\bibitem{aceituno2017simultaneous}
B.~Aceituno-Cabezas, C.~Mastalli, H.~Dai, M.~Focchi, A.~Radulescu, D.~G.
  Caldwell, J.~Cappelletto, J.~C. Grieco, G.~Fern{\'a}ndez-L{\'o}pez, and
  C.~Semini, ``Simultaneous contact, gait, and motion planning for robust
  multilegged locomotion via mixed-integer convex optimization,'' {\em IEEE
  Robotics and Automation Letters}, vol.~3, no.~3, pp.~2531--2538, 2017.

\bibitem{ding2020kinodynamic}
Y.~Ding, C.~Li, and H.-W. Park, ``Kinodynamic motion planning for multi-legged
  robot jumping via mixed-integer convex program,'' in {\em IEEE/RSJ
  International Conference on Intelligent Robots and Systems (IROS)}, 2020.

\bibitem{mccormick1976computability}
G.~P. McCormick, ``Computability of global solutions to factorable nonconvex
  programs: Part i—convex underestimating problems,'' {\em Mathematical
  programming}, vol.~10, no.~1, pp.~147--175, 1976.

\bibitem{shkolnik2011bounding}
A.~Shkolnik, M.~Levashov, I.~R. Manchester, and R.~Tedrake, ``Bounding on rough
  terrain with the littledog robot,'' {\em The International Journal of
  Robotics Research}, vol.~30, no.~2, pp.~192--215, 2011.

\bibitem{manchester2011stable}
I.~R. Manchester, U.~Mettin, F.~Iida, and R.~Tedrake, ``Stable dynamic walking
  over uneven terrain,'' {\em The International Journal of Robotics Research},
  vol.~30, no.~3, pp.~265--279, 2011.

\bibitem{adabag2023mpcgpu}
E.~Adabag, M.~Atal, W.~Gerard, and B.~Plancher, ``{MPCGPU: Real-Time Nonlinear
  Model Predictive Control through Preconditioned Conjugate Gradient on the
  GPU},'' in {\em IEEE International Conference on Robotics and Automation
  (ICRA)}, 2024.

\bibitem{jeon2024cusadi}
S.~H. Jeon, S.~Hong, H.~J. Lee, C.~Khazoom, and S.~Kim, ``Cusadi: A gpu
  parallelization framework for symbolic expressions and optimal control,''
  {\em arXiv preprint arXiv:2408.09662}, 2024.

\bibitem{wachter2006implementation}
A.~W{\"a}chter and L.~T. Biegler, ``On the implementation of an interior-point
  filter line-search algorithm for large-scale nonlinear programming,'' {\em
  Mathematical programming}, vol.~106, pp.~25--57, 2006.

\bibitem{chatzilygeroudis2018using}
K.~Chatzilygeroudis and J.-B. Mouret, ``Using parameterized black-box priors to
  scale up model-based policy search for robotics,'' in {\em IEEE International
  Conference on Robotics and Automation (ICRA)}, 2018.

\end{thebibliography}

\end{document}